\documentclass[10pt,twocolumn,letterpaper]{article}

\usepackage{cvpr}              

\definecolor{cvprblue}{rgb}{0.21,0.49,0.74}
\usepackage[pagebackref,breaklinks,colorlinks,allcolors=cvprblue]{hyperref}
\usepackage{algorithm}
\usepackage{algorithmic}

\usepackage{amsmath}
\usepackage{amsthm}
\usepackage{amsfonts}
\usepackage{colortbl}
\usepackage{booktabs} 
\usepackage{xcolor}
\usepackage{subcaption}
\usepackage{caption}
\usepackage{subcaption}
\usepackage{multirow}
\usepackage{lipsum} 
\usepackage{multicol} 
\usepackage{mathrsfs}

\newtheorem{Definition}{Definition}
\newtheorem{Theorem}{Theorem}

\newcommand{\argmax}{\mathop{\mathrm{argmax}}}   

\def\paperID{7621} 
\def\confName{CVPR}
\def\confYear{2025}

\title{Detecting Adversarial Data Using Perturbation Forgery}

\author {
    Qian Wang\textsuperscript{\rm 1}, 
    Chen Li\textsuperscript{\rm 1}, \
    Yuchen Luo\textsuperscript{\rm 2}, \
    Hefei Ling\textsuperscript{\rm 1}\thanks{Corresponding author}, \
    Shijuan Huang\textsuperscript{\rm 1}, \
    Ruoxi Jia\textsuperscript{\rm 3}, \
    Ning Yu\textsuperscript{\rm 4} \\
    \textsuperscript{\rm 1}Huazhong University of Science and Technology, China \quad
    \textsuperscript{\rm 2}Wuhan University, China\\
    \textsuperscript{\rm 3}Virginia Tech, USA \quad
    \textsuperscript{\rm 4}Netflix Eyeline, USA\\
    \{yqwq1996, lhefei, llichen, shijuan\_huang\}@hust.edu.cn, \
    \{yuchenluo\}@whu.edu.cn, \\
    ruoxijia@vt.edu, \
    ningyu.hust@gmail.com
}

\begin{document}
\maketitle

\begin{abstract}
As a defense strategy against adversarial attacks, adversarial detection aims to identify and filter out adversarial data from the data flow based on discrepancies in distribution and noise patterns between natural and adversarial data. 
Although previous detection methods achieve high performance in detecting gradient-based adversarial attacks, new attacks based on generative models with imbalanced and anisotropic noise patterns evade detection. 
Even worse, the significant inference time overhead and limited performance
against unseen attacks make existing techniques impractical for real-world use.
In this paper, we explore the proximity relationship among adversarial noise distributions and demonstrate the existence of an open covering for these distributions. 
By training on the open covering of adversarial noise distributions, a detector with strong generalization performance against various types of unseen attacks can be developed.
Based on this insight, we heuristically propose Perturbation Forgery, which includes noise distribution perturbation, sparse mask generation, and pseudo-adversarial data production, to train an adversarial detector capable of detecting any unseen gradient-based, generative-based, and physical adversarial attacks.
Comprehensive experiments conducted on multiple general and facial datasets, with a wide spectrum of attacks, validate the strong generalization of our method.\footnote{Code at \href{https://github.com/cc13qq/PFD}{https://github.com/cc13qq/PFD}}
\end{abstract}

%

\begin{table*}[ht]
    \centering
    \tabcolsep=0.15cm
    \caption{
     \textbf{
     Properties of our method vs Other Adversarial Detection}: We summarize four metrics for adversarial detection.
     “Detect Attacks” specifies the types of attacks each method is designed to detect. 
      “Model-Agnostic” indicates whether the detection method operates independently of the protected model. 
      “Unseen-Attack Detection” assesses the method’s generalization to detect previously unseen attacks, with the reported value representing the AUROC score for detecting the MIFGSM attack on the ImageNet dataset. 
      Lastly, “Time Overhead” quantifies the time required for the detector to process 100 samples from the ImageNet dataset.
    } 
    \begin{tabular}{c|lccc}
        \toprule
         \textbf{Methods} & \textbf{Detect Attacks} & \textbf{Model-Agnostic} & \textbf{Unseen-Attack Detection} & \textbf{Time Overhead} \\
        \midrule
         LID~\cite{LID} & Gradient & $\times$ & 0.9146 & 1.80s  \\ 
         LiBRe~\cite{Libre} & Gradient & $\times$ & 0.8725 & 2.56s\\ 
         SPAD~\cite{Wang:23selfperturbation} & Gradient + GAN & \checkmark & 0.9820 & 4.56s \\
         EPSAD~\cite{EPSAD} & Gradient & \checkmark & 0.9918  & 396.81s  \\
        \midrule
        \textbf{Ours} & Gradient + GAN + Diffusion & \checkmark & 0.9931 & 4.85 \\
        \bottomrule
    \end{tabular}
    
    \label{tab: baseline}
\end{table*}

\section{Introduction}
\label{Sec: Introduction}
Numerous studies have demonstrated the effectiveness of deep neural networks (DNNs) in various tasks~\cite{Shi:20Learning}. 
However, it is also well-known that DNNs are vulnerable to adversarial attacks, which generate adversarial data by adding imperceptible perturbations to natural images, potentially causing the model to make abnormal predictions. 
This vulnerability reduces the reliability of deep learning systems, highlighting the urgent need for defense techniques.
Some approaches employ adversarial training techniques~\cite{TRADES}, which incorporate adversarial data into the training process to inherently bolster the model's immunity against adversarial noise (adv-noise). 
Other methods utilize adversarial purification~\cite{ADP} to cleanse the input data of adv-noise before model inference.
Unfortunately, adversarial training often results in diminished classification accuracy on adversarial data and may sacrifice performance on natural data~\cite{CircumDefense}. 
While purification techniques may seem more reliable, the denoising process can inadvertently smooth the high-frequency texture of clean images, leading to suboptimal accuracy on both natural and adversarial images.

Another branch of adversarial defense is adversarial detection~\cite{KD, LID}, which aims to filter out adversarial data before they are fed into target systems. 
These methods rely on discerning the discrepancies between adversarial and natural distributions, thereby maintaining the integrity of target systems while achieving high performance in defending against adversarial attacks.
However, existing adversarial detection approaches primarily train specialized detectors tailored for specific attacks or classifiers~\cite{Libre}, which may not generalize well to unseen advanced attacks~\cite{DetectHard,EvadDetect}.

Some recent approaches~\cite{EPSAD, Wang:23selfperturbation} have attempted to enhance the generalization of adversarial detectors by leveraging diffusion models and manually designed pseudo-noise. 
Nevertheless, as shown in Table~\ref{tab: baseline}, this improvement often comes at the expense of increased inference overhead, and even well-trained detectors still struggle with adversarial data produced by generative models such as generative adversarial networks (GANs) and diffusion models.
Specifically, generative-based adversarial attacks, such as M3D~\cite{M3D} and Diff-PGD~\cite{Diff-pgd}, are recently developed techniques designed to generate more natural adversarial data. 
Unlike traditional gradient-based attacks like PGD~\cite{PGD}, generative-based attacks tend to create imbalanced and anisotropic perturbations that are intense in high-frequency and salient areas but mild in low-frequency and background areas. 
These perturbations make them harder to detect with current methods, posing a new challenge to the field of adversarial detection.
Therefore, there is an urgent need for a low-time-cost detection method with strong generalization performance against both unseen gradient-based attacks and unseen generative-based attacks.

\begin{figure*}[ht]
    \centering
    \includegraphics[width=2\columnwidth]{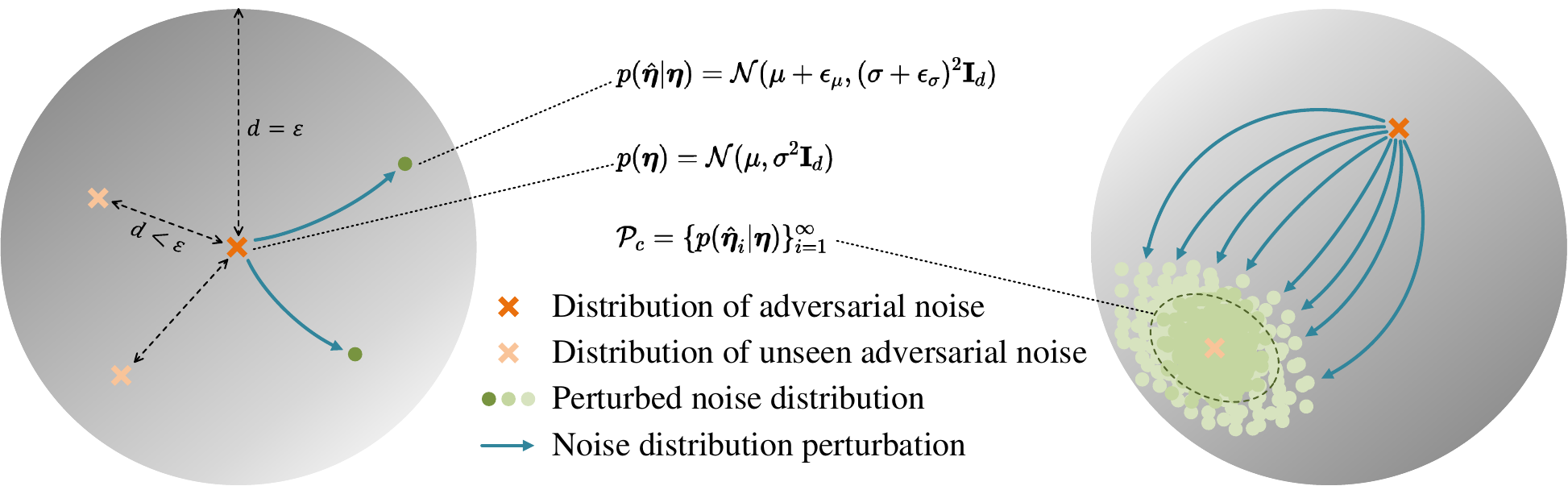}
    \caption{
    Illustration of the proximity of adversarial noise distributions and perturbed noise distributions.
    Left: all the distributions of adversarial noise and the perturbed distributions are in a $\varepsilon$-ball centered on a given adversarial noise distribution.
    Right: by continuously perturbing the given adversarial noise distribution, we obtain an open covering of distributions of adversarial noise.
    }
    \label{fig: proximity}
\end{figure*}

In this paper, we investigate the association between the distributions of different adv-noises. 
As shown in Figure~\ref{fig: proximity}, by modeling noise as multivariate Gaussian distributions, we define the proximity relationship of the noise distributions in a metric space and establish a perturbation method to construct the proximal distributions.
Based on the assumption and definition, we demonstrate that all adv-noise distributions are proximal using the Wasserstein distance, and we deduce a corollary that an open covering of the adv-noise distributions exists.
We then propose the core idea of this paper: 
\textbf{
by training on the open covering, we obtain a detector with strong generalization performance against various types of unseen attacks.
}

Based on this idea, we propose Perturbation Forgery, a pseudo-adversarial data (pseudo-adv-data) generation method designed to train a detector with strong generalization capabilities for detecting various types of adversarial attacks. 
As shown in Figure~\ref{fig: pipeline}, Perturbation Forgery comprises noise distribution perturbation, sparse mask generation, and pseudo-adversarial data (pseudo-adv-data) production.
By continuously perturbing the noise distribution of a commonly used attack, such as FGSM~\cite{FGSM}, we obtain a nearly complete family of distributions that form the aforementioned open covering. 
Simultaneously, sparse masks are generated using attention maps and saliency maps of the natural data, converting the sampled global noises into local forms to help the detector focus on local noise patterns so as to detect generative-based adv-noise.
By sampling noise from the perturbed noise distributions, we sparsify it using sparse masks and inject it into natural data. 
This process converts half of the natural data into pseudo-adv-data, replacing actual adversarial data. 
The pseudo-adv-data is then combined with the remaining uncontaminated data to train the binary adversarial detector.

To the best of our knowledge, we are the first to empower a detector with the robust ability to detect various types of adversarial attacks, including gradient-based, GAN-based, diffusion-based, and physical attacks, with minor inference time overhead.

Our contributions are summarized in four thrusts:

$\bullet$ By modeling adversarial noise as Gaussian distributions, we investigate the association between different adversarial noise distributions and demonstrate their proximity in a metric space.

$\bullet$ Based on our assumption and definition, we deduce the existence of an open covering for adversarial noise distributions and propose training a detector to distinguish this open covering from the natural data distribution, thereby achieving strong generalization in detection.

$\bullet$ Building on our deductions, we propose Perturbation Forgery, which includes noise distribution perturbation, sparse mask generation, and pseudo-adversarial data production. 
This method is designed to train a binary detector with strong generalization performance against various types of unseen attacks, including gradient-based, GAN-based, diffusion-based, and physical adversarial data, with minor inference time overhead.

$\bullet$ We conduct evaluations on multiple general and facial datasets across a wide spectrum of gradient-based, generative-based, and physical attacks, demonstrating the consistently strong performance of our method in adversarial detection tasks.

\section{Related Work}
\label{Sec: Related work}

\subsection{Adversarial Attack}
\label{Adversarial attacks}
\paragraph{Gradient-based adversarial attack}
Adversarial attacks~\cite{FGSM} attempt to fool a classifier into changing prediction results by injecting subtle perturbations to the input data while maintaining imperceptibility from human eyes.
By introducing iterative perturbation, PGD~\cite{PGD} and BIM~\cite{BIM} generate adversarial data with significantly improved attack performance, while DIM~\cite{DIFGSM}, NIM~\cite{NIFGSM_SNIM}, and MIM~\cite{MIFGSM} improve the transferability under the black-box setting.
Besides, adaptive attacks~\cite{AutoAttack,MM_Attack,BPDA_EBM} are proposed to exploit the weak points of each defense, achieving a high attack rate against several defense techniques. 

\paragraph{Generative-based Adversarial Attacks}
Recently, another branch of attacks uses generative models to produce adversarial data with a more natural style and imbalanced noise compared to gradient-based attacks.
Among them, CDA~\cite{CDA}, TTP~\cite{TTP}, and M3D~\cite{M3D} utilize GAN to craft adversarial data and achieve higher efficiency when attacking large-scale datasets.
Diff-Attack~\cite{Diff-attack} and Diff-PGD~\cite{Diff-pgd} employ the diffusion model as a strong prior to better ensure the realism of generated data.
Designed for perturbing facial recognition systems, AdvMakeup~\cite{AdvMakeup} and AMT-GAN~\cite{AMT-GAN} are presented to generate imperceptible adversarial images of the face.

In this paper, a wide spectrum of gradient-based, generative-based, and physical adversarial attacks on ImageNet and facial datasets is used to  evaluate the detection performance of our method.



\subsection{Adversarial Detection}
\label{Sec: Adversarial detection}
As one of the technical solutions to ensure the safety of DNNs, there has been a significant amount of exploration in the field of adversarial detection~\cite{LID}.
LiBRe~\cite{Libre} leverages Bayesian neural networks to endow pre-trained DNNs with the ability to defend against unseen adversarial attacks, SPAD~\cite{Wang:23selfperturbation} leverages a data augmentation method to detect adversarial face images from unseen attack, while EPSAD~\cite{EPSAD} incorporates a diffusion model to detect adversarial data with minimal attack intensity, albeit at the cost of inference time.

While current methods have achieved remarkable improvements in adversarial detection, generative-based adversarial attacks are harder to detect, posing a new challenge.
To address the problems above, our method is proposed to detect various types of unseen attacks, including gradient-based, generative-based, and physical attacks.

\section{Proximity of Adversarial Noise Distribution}
\label{Sec: Proximity of adv-noise}
In this section, we provide an overview of preliminary concepts related to adversarial data generation and explore the relationship between adversarial noise distributions.
We define the proximity of noise distributions and present the core idea of this paper: by training on the open covering of adversarial noise distributions, we obtain a detector with strong generalization performance against various types of unseen attacks.


\subsection{Adversarial data generation}
Given a DNN $f$ that works on dataset $\mathcal{D} = \{(\mathbf{x}_i, y_i)\}^n_{i=1}$ with $(\mathbf{x}_i \in \mathcal{X}, y_i \in \mathcal{C})$ being (sample, ground-truth) pair, an adversarial example $\hat{\mathbf{x}}$ regarding sample $\mathbf{x}$ with attack intensity $\epsilon$ is calculated as:
\begin{equation}
\label{Eq: adv-attack}
    \hat{\mathbf{x}} = \argmax_{\hat{\mathbf{x}}\in \mathcal{B}(\mathbf{x}, \epsilon)} \mathcal{L}(f(\hat{\mathbf{x}}), y),
\end{equation}
where $\mathcal{L}$ is some loss function and $\mathcal{B}(\mathbf{x}, \epsilon) = \{\mathbf{x'} | \ell^p(\mathbf{x}, \mathbf{x'}) < \epsilon \}$ restricts the perturbation under $\ell^p$ norm. 
Following~\cite{FGSM}, we simply denote adversarial sample as $\hat{\mathbf{x}} = \mathbf{x} + \boldsymbol{\eta}$.

\begin{figure*}[ht]
    \centering
    \includegraphics[width=2.1\columnwidth]{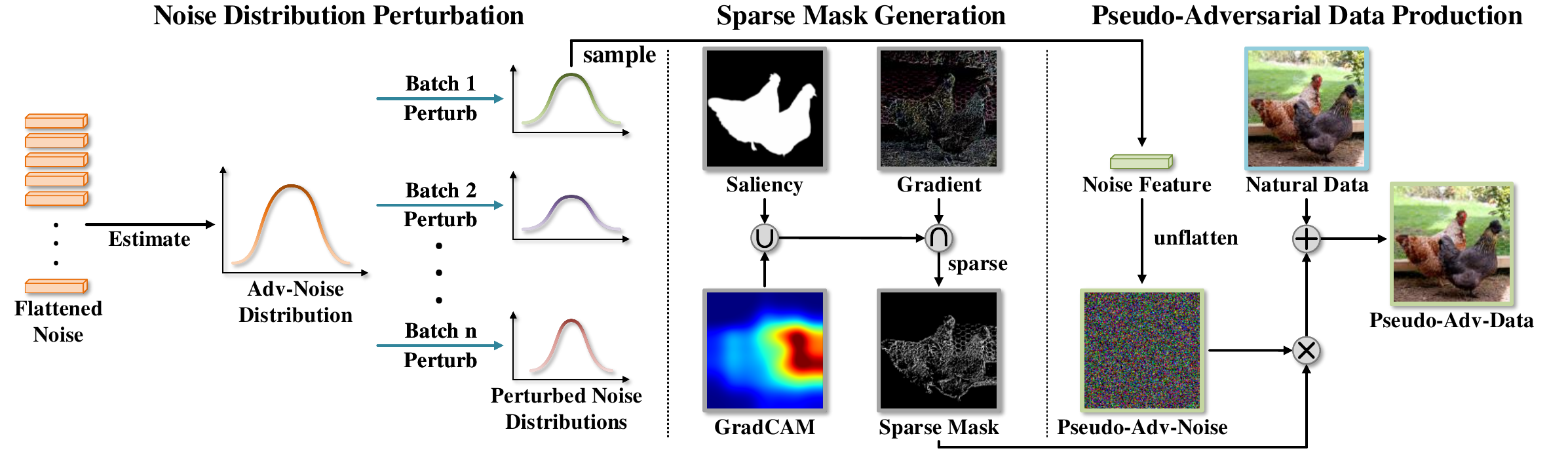}
    \caption{
    Illustration of Perturbation Forgery.
    Before training, we estimate the noise distribution from a commonly used attack, then continuously perturb it in each batch to create an open covering of the adversarial noise distributions. 
    Next, noises sampled from these perturbed distributions are converted into localized noise by applying sparse masks. 
    Finally, pseudo-adversarial data is generated by adding these localized noises to natural samples.
    }
    \label{fig: pipeline}
\end{figure*}

\subsection{Proximity of noise distribution}

Due to the subtle and imperceptible nature of adv-noise, adversarial data and natural data appear visually similar. 
Thus, we assume that this similarity also exists between the noises of different adversarial attacks. 
To investigate the similarity of these noises, we provide the following definition, which depicts a close relationship between two noise distributions.
\begin{Definition}[\textbf{Proximal Noise Distribution}]
\label{Definition: similarity}
Let $P$ and $Q$ be the distribution functions of two independent noise samples $\mathbf{x}_1$ and $\mathbf{x}_2$. 
We call them proximal noise distributions under $\varepsilon$ if we have a metric $d(\cdot, \cdot)$ and $\exists \ \varepsilon > 0$, s.t. $d(P, Q) < \varepsilon$.
\end{Definition}

A common assumption for instance feature distribution is to model it as a Gaussian distribution~\cite{VOS}.
Obtained by subtracting from natural data and adversarial data, adv-noises naturally form a Gaussian distribution.
Therefore, we assume that noise r.v $\boldsymbol{\eta}\in \mathbb{R}^d$ satisfied  truncated Gaussian distribution $\mathcal{N}(\boldsymbol{\mu}, \sigma^{2} \mathbf{I}_{d})$ with it's truncated interval is $[-\epsilon, \epsilon]^d$. 
Built upon Definition~\ref{Definition: similarity}, we derive the following theorem about the association between the distributions of different adv-noises:

\begin{Theorem}
\label{Theorem: similarity}
Let $\mathcal{P}_a$ be the distribution set composed of all the adversarial noise distributions.
Given independent noise distributions $Q_i, i \in N^+$.
For $\forall \ i \neq j$, $Q_i$ and $Q_j$ are proximal noise distributions if the following conditions are met.

1) $Q_i, Q_j \in \mathcal{P}_a$.

2) $\exists \ \epsilon_{\mu}, \epsilon_{\sigma} > 0$ s.t. $\|\boldsymbol{\mu}_i - \boldsymbol{\mu}_j\| < \epsilon_{\mu}$ and $|\sigma_i - \sigma_j| < \epsilon_{\sigma}$ where the $\| \cdot \|$ is a Euclidean norm on $\mathbb{R}^n$.
\end{Theorem}


Theorem~\ref{Theorem: similarity} tells us:
1) All the adversarial noise distributions under the same $\ell^p$ norm are proximal noise distributions.
2) If the parameters of two noise distributions are sufficiently close to those of a given distribution, we can regard these two distributions as proximal noise distributions.

We can further deduce several corollaries of the theorem.
1) All noise distributions proximal to the known adversarial distribution $p(\boldsymbol{\eta})$ form an open $\varepsilon$-ball centered on $p(\boldsymbol{\eta})$, denoting as $\mathcal{P}_{\varepsilon} = \{p(\boldsymbol{\eta}_i) | d(p(\boldsymbol{\eta}_i), p(\boldsymbol{\eta})) < \varepsilon) \}$, and also form a metric space $(\mathcal{P}_{\varepsilon}, d)$.
2) The adv-noise distribution set is also located in this ball: $\mathcal{P}_{a} \subset \mathcal{P}_{\varepsilon}$.
3) Based on the properties of the metric space $(\mathcal{P}_{\varepsilon}, d)$, for all subset $\mathcal{P}_{i} \subset \mathcal{P}_{\varepsilon}$, there exists an open covering of the subset: $\mathcal{P}_{i} \subset \{G_c\}_{c\in \mathbb{I}}$.

Based on the corollaries, we deduce the core idea of this paper:
Let $p(\hat{\boldsymbol{\eta}}_i|\boldsymbol{\eta}) = \mathcal{N}(\boldsymbol{\mu} + \epsilon_{\mu i}, (\sigma + \epsilon_{\sigma i})^2 \mathbf{I}_{d}), i = 1, 2, \dots, n$ denote generated proximal noise distributions of the given adv-noise distribution $p(\boldsymbol{\eta}) = \mathcal{N}(\boldsymbol{\mu}, \sigma^2 \mathbf{I}_{d})$, where $\epsilon_{\mu i}, \epsilon_{\sigma i}$ are randomly sampled from uniform distributions and restricted by $\epsilon_{\mu i}, \epsilon_{\sigma i} < \varepsilon$.
As shown in Figure~\ref{fig: proximity}, \textbf{when $n \rightarrow \infty$, we can always obtain a distribution set $\mathcal{P}_c = \{ p(\hat{\boldsymbol{\eta}}_i|\boldsymbol{\eta}) \}_{i=1}^\infty$ that forms an open covering of the adversarial distribution set $\mathcal{P}_a$.} 
\textbf{By training on the open covering, the detector learns to distinguish between the distributions of natural and adversarial data, enabling it to effectively detect unseen adversarial attacks.}

Analysis of distribution assumption, proofs of theorem, and corollaries are provided in the supplementary materials.

\section{Perturbation Forgery}
\label{Sec: Perturbation Forgery}
Based on the theory in the last section, we heuristically propose Perturbation Forgery to train a detector with strong generalization against various types of unseen attacks.

As illustrated in Figure~\ref{fig: pipeline}, Perturbation Forgery encompasses three elements: noise distribution perturbation, sparse mask generation, and pseudo-adversarial data production.
At first, the noise distribution of a commonly used attack is continuously perturbed using noise distribution perturbation into numerous noise distributions that remain proximal to the original, thereby forming the open covering.
After that, we sample noises from the perturbed distribution and convert them into sparse local noise using sparse mask generation. 
Subsequently, we convert half of the natural data into pseudo-adv-data using pseudo-adversarial data production.
This pseudo-adv data replaces actual adversarial data and is combined with the remaining natural data to train a binary detector.

\subsection{Noise Distribution Perturbation}
\label{Sec: Noise Distribution Perturbation}

Given the training dataset $\{(\mathbf{x}_k, y_k)\}^n_{k=1}$ composed by natural data, we add adv-noise on it using a commonly used attack (such as FGSM) and generate adversarial noise features $\mathcal{D}_{\mathrm{noise}} := \{\mathbf{z}_k = h(\boldsymbol{\eta}_k) \}_{k=1}^n$ by
\begin{equation}
\label{Eq: adv-noise generation}
    \boldsymbol{\eta}_k = \operatorname{Attack} (\mathbf{x}) - \mathbf{x}, \ k \in [1, n],
\end{equation}
where $h: \mathcal{X} \rightarrow \mathbb{R}^d$ is the flattening operation that converts multidimensional data into vectors.

Assuming the noise features form a multivariate Gaussian distribution $p(\mathbf{z}) = \mathcal{N}(\boldsymbol{\mu}, \boldsymbol{\Sigma})$~\cite{VOS}, 
we estimate the parameters mean $\boldsymbol{\mu}$ and covariance $\boldsymbol{\Sigma}$ of $\mathcal{D}_{\mathrm{noise}}$ by
\begin{equation}
\label{Eq: distribution estimation}
    \widehat{\boldsymbol{\mu}} =\frac{1}{N} \sum_{k} \mathbf{z}_k, \quad 
    \widehat{\boldsymbol{\Sigma}} =\frac{1}{N} \sum_{k}\left( \mathbf{z}_k -\widehat{\boldsymbol{\mu}}\right)\left( \mathbf{z}_k - \widehat{\boldsymbol{\mu}}\right)^{\top}.
\end{equation}

For the $i^{\mathrm{th}}$ batch of training, we perturb the estimated distribution to a proximal noise distribution by
\begin{equation}
\label{Eq: distribution perturbation}
    \widehat{\boldsymbol{\mu}}_i = \widehat{\boldsymbol{\mu}} + \alpha_i \cdot \boldsymbol{m}_i, \quad 
    \widehat{\boldsymbol{\Sigma}}_i = \widehat{\boldsymbol{\Sigma}} + \beta_i \cdot \boldsymbol{V}_i,
\end{equation}
where $\boldsymbol{m}_i \sim \mathcal{N}(0, \mathbf{I}_{d})$ and $\boldsymbol{V}_i \sim \mathcal{N}(0, \mathbf{I}_{d \times d})$ are vector and matrix of perturbation. 
$\alpha_i \sim \boldsymbol{U}(-\epsilon_{\mu}, \epsilon_{\mu})$ and $\beta_i \sim \boldsymbol{U}(0, \epsilon_{\sigma})$ are scale parameters.
After $n$ batches of training, we have a perturbed distribution set $\{\mathcal{N}(\widehat{\boldsymbol{\mu}}_i, \widehat{\boldsymbol{\Sigma}}_i) \}$, thereby forming the approximate open covering $\mathcal{P}_c$. 
Trained with $\mathcal{P}_c$, the detector can distinguish the distributions of natural and adversarial data. 

\subsection{Sparse Mask Generation}
\label{Sec: Sparse Mask Generation}

Unlike gradient-based adversarial attacks, generative-based attacks tend to generate imbalanced and anisotropic perturbations, showing much intensity in high-frequency and salient areas and mild in low-frequency and background areas. 
Besides, in physical scenarios, adversarial attacks also tend to add patches to local areas.
Noticing this, we propose sparse masks to perturbation forgery to assist the detector in detecting local adv-noises.

Previous research has demonstrated that the spectra of real and fake images differ significantly in high-frequency regions~\cite{YuchenLuo2021}. 
Additionally, generative-based adversarial data manipulate characteristics, such as abnormal color aberrations, in these high-frequency regions~\cite{Wang:23selfperturbation}.
To augment the detection performance on generative-based adversarial data, we need to sparse the pseudo-noise sampled from perturbed noise distribution and highlight the noise in high-frequency regions to mimic the style of generative-based attacks.

\begin{algorithm}[ht]
\caption{Training Procedure with Perturbation Forgery}
\label{alg: Perturbation Forgery}
\textbf{Input}: Training dataset, a commonly used attack, saliency detection model $f_s$, and GradCAM model $f_c$.\\
\textbf{Output}: A detector with strong generalization capabilities against various types of unseen attacks.

\begin{algorithmic}[1] 
\STATE Generate adversarial noise using the commonly used attack in Eq.~\ref{Eq: adv-noise generation}.
\STATE Estimate a multivariate Gaussian distribution $p(\mathbf{z}) = \mathcal{N}(\widehat{\boldsymbol{\mu}}, \widehat{\boldsymbol{\Sigma}})$ of adv-noise using Eq.~\ref{Eq: distribution estimation}.
\FOR{batch $i$ \textbf{in} $1,\cdots ,n$}
\STATE Generate a perturbed noise distribution $p(\mathbf{z}_i) = \mathcal{N}(\widehat{\boldsymbol{\mu}}_i, \widehat{\boldsymbol{\Sigma}}_i)$ using Eq.~\ref{Eq: distribution perturbation}.
\FOR{sample $\mathbf{x}_{i,k}$ in half of batch $i$}
\STATE Generate sparse masks $\operatorname{Mask}(\mathbf{x}_{i,k})$ using Eq.~\ref{Eq: sparse mask1}.
\STATE Sample pseudo-adv noise $\widehat{\boldsymbol{\eta}}_{i,k}$ using Eq.~\ref{Eq: sample pseudo noise}.
\STATE Generate pseudo-adv data $\widehat{\mathbf{x}}_{i,k}$ using Eq.~\ref{Eq: pseudo-adv-sample generation} and replace $\mathbf{x}_{i,k}$.
\ENDFOR
\STATE Train the detector on batch $i$.
\ENDFOR
\end{algorithmic}
\end{algorithm}

Given this, for a natural sample $\mathbf{x}_{i,k}$ in the $\text{i}^{th}$ batch, we first mask the high-frequency area using saliency detection~\cite{saliency} and GradCAM~\cite{GradCAM}:
\begin{equation}
\label{Eq: high-frequency mask}
    \operatorname{Mask}_1(\mathbf{x}_{i,k}) = \operatorname{Map}(f_s(\mathbf{x}_{i,k}); \gamma_{s}) \cup \operatorname{Map}(f_c(\mathbf{x}_{i,k}); \gamma_{c}),
\end{equation}
where $f_s$ and $f_c$ are the saliency detection model and GradCAM model respectively. $\operatorname{Map}(\mathbf{x}; \gamma)$ is a binary indicator function parameterized by $\gamma$:
\begin{equation}
    \operatorname{Map}(\mathbf{x}_{i,k}; \gamma) = \operatorname{Clip}_{[0,1]} (\mathbf{x}_{i,k} - \gamma \mathbf{I}_{\mathcal{X}}).
\end{equation}

The adv-noise that occurs in the high-frequency region tends to occupy only a small part of the region. 
Therefore, we select the high-frequency point from the gradient map obtained by Sobel operator~\cite{SobelOperator} to sparsify the mask of sample $\mathbf{x}_{i,k}$:
\begin{align}
\label{Eq: sparse mask}
    &\operatorname{Mask}_2(\mathbf{x}_{i,k}) = \operatorname{Map}(\operatorname{Sobel}(\mathbf{x}_{i,k}); \gamma_{l}), \\ 
    &\operatorname{Mask}(\mathbf{x}_{i,k}) = \operatorname{Mask}_1(\mathbf{x}_{i,k}) \cap \operatorname{Mask}_2(\mathbf{x}_{i,k}).
    \label{Eq: sparse mask1}
\end{align}

\subsection{Pseudo-Adversarial Data Production}
\label{Sec: Pseudo-adv-sample Genration}

For the $i^{\mathrm{th}}$ batch of training, we sample pseudo-adv noise from the perturbed distribution by:
\begin{align}
\label{Eq: sample pseudo noise}
    &\widehat{\boldsymbol{\eta}}_{i,k} = h^{-1}(\mathbf{z}_{i,k}), \\
    &\mathbf{z}_{i,k} \in \{\mathbf{z}_k | \mathbf{z}_{k} \sim \mathcal{N}(\widehat{\boldsymbol{\mu}}_i, \widehat{\boldsymbol{\Sigma}}_i) \ \mathrm{and} \ p(\mathbf{z}_{k}) < \gamma_{p}    \}
\end{align}
where $h^{-1}: \mathbb{R}^d \rightarrow \mathcal{X}$ is an inverse function of $h$ to unflatten noise features, and $\gamma_p$ is a threshold that excludes data with a low likelihood.
Note that $\boldsymbol{\eta}_{i}$ is a global noise and needs to be localized by a sparse mask.
The pseudo-adv data $\widehat{\mathbf{x}}_{i,k}$ is produced by adding the masked local noise to $\mathbf{x}_{i,k}$:

\begin{equation}
\label{Eq: pseudo-adv-sample generation}
    \widehat{\mathbf{x}}_{i,k} = \mathbf{x}_{i,k} +  \widehat{\boldsymbol{\eta}}_{i,k} \otimes \operatorname{Mask}(\mathbf{\mathbf{x}_{i,k}}).
\end{equation}
where $\otimes$ is Hadamard product.

For each batch, we convert half of the natural data to pseudo-adv data by Perturbation Forgery and train a binary detector using the Cross-Entropy classification loss~\cite{CrossEntropyLoss}.
The overall training procedure with Perturbation Forgery is presented in Algorithm~\ref{alg: Perturbation Forgery}.

\begin{table*}[ht]
\footnotesize
    \centering
    \tabcolsep=0.4cm
    \caption{
    Comparison of AUROC scores of detecting gradient-based and adaptive attacks on ImageNet100 under $\epsilon=4/255$. 
    }
    \begin{tabular}{l|ccccccccccc}
        \toprule
        Detector & BIM & PGD & DIM & MIM & NIM & AA & BPDA+EOT & MM\\
        \hline
        LID~\cite{LID} & 0.9782 & 0.9750 & 0.8942 & 0.9146 & 0.8977 & 0.9124 & 0.9202 & 0.9227\\
        LiBRe~\cite{Libre} & 0.9259 & 0.9548 & 0.9243 & 0.8725 & 0.9013 & 0.8653 &0.8714 & 0.8573\\
        SPAD~\cite{Wang:23selfperturbation} & 0.9846 & 0.9851 & 0.9815 & 0.9820 & 0.9823 & 0.9890 & 0.9885 & 0.9811  \\
        
        EPSAD~\cite{EPSAD} & \textbf{0.9998} & \textbf{0.9989} & \textbf{0.9923} & 0.9918 & \textbf{0.9972} & \textbf{0.9998} & \textbf{0.9985} & 0.9923\\
        
        \hline
        \textbf{ours} & 0.9911 & 0.9912 & 0.9863 & \textbf{0.9931} & 0.9934 & 0.9941 & 0.9927 & \textbf{0.9944}\\
        \bottomrule
    \end{tabular}
    \label{Tab: ImageNet100-gradient}
\end{table*}

\section{Experiments}
\label{Sec: Experiments}

To validate the detection performance of our approach against various adversarial attack methods, we conduct extensive experiments on multiple general and facial datasets. 
For additional experiments, analyses, and a discussion of limitations, please refer to the supplementary material.

\subsection{Experiment Settings}
\label{Sec: Experiment Setting}

\paragraph{Datasets}
The datasets used for evaluation include: 1) general datasets: ImageNet100~\cite{ImageNet} and CIFAR-10~\cite{CIFAR10} for evaluate gradient-based and generative-based attacks, and 2) face datasets: Makeup~\cite{Makeup} for Adv-Makeup~\cite{AdvMakeup}, CelebA-HQ~\cite{CelebA-HQ} for AMT-GAN~\cite{AMT-GAN}, and LFW~\cite{LFW} for TIPIM~\cite{TIPIM} and physical attacks.

\begin{table*}[ht]
\footnotesize
    \centering
    \tabcolsep=0.616cm
    \caption{
    Comparison of AUROC scores of detecting generative-based attacks on ImageNet100. 
    }
    \begin{tabular}{l|ccc|cc}
        \toprule
        Detector & CDA & TTP & M3D & Diff-Attack & Diff-PGD  \\
        \hline
        CNN-Detection~\cite{cnn-detection}
        & 0.7051 & 0.6743 & 0.6917 & 0.3963 & 0.5260 \\
        LGrad~\cite{LGRAD}
        & 0.6144 & 0.6077 & 0.6068 & 0.5586 & 0.5835 \\
        Universal-Detector~\cite{universal-detector}
        & 0.7945 & 0.8170 & 0.8312 & 0.9202 & 0.5531 \\
        DIRE~\cite{dire}
        & 0.8976 & 0.9097 & 0.9129 & 0.9097 & 0.9143 \\
        \hline
        SPAD
        & 0.9385 & 0.9064 & 0.8984 & 0.8862 & 0.8879 \\
        EPSAD
        & 0.9674 & 0.6997 & 0.7265 & 0.4700 & 0.1507 \\
        \hline
        \textbf{ours} & \textbf{0.9878} & \textbf{0.9678} & \textbf{0.9364} & \textbf{0.9223} & \textbf{0.9223} \\
        \bottomrule
    \end{tabular}
    \label{Tab: ImageNet100-generative}
\end{table*}
        
\begin{table*}[ht]
\footnotesize
    \centering
    \tabcolsep=0.25cm
    \caption{
    Comparison of AUROC scores for detecting various adversarial attacks on the face dataset. 
    }
    \begin{tabular}{l|c|ccc||lcc}
        \toprule
        Detector & TIPIM & Adv-Sticker & Adv-Glasses & Adv-Mask & Detector & Adv-Makeup & AMT-GAN   \\
        \hline
        LID
        &  0.6974 & 0.5103 & 0.6472 & 0.5319
        & He et al.~\cite{YangHe2021} & 0.5252 & 0.8823 \\
        LiBRe
        & 0.5345 & 0.9739 &0.8140  & 0.9993
        & Luo et al.~\cite{YuchenLuo2021} & 0.6178 & 0.6511 \\
        EPSAD
        & 0.9647 & 0.9462 & 0.9260 & 0.9578 
        & ODIN\cite{ODIN} & 0.6325 & 0.6970 \\
        SPAD
        & 0.9256  & 0.9975 & 0.9244 & 0.9863 
        & SPAD & 0.9657 & 0.8969 \\
        \hline
        \textbf{ours} & \textbf{0.9885} & \textbf{0.9999} & \textbf{0.9999} & \textbf{0.9999} 
        & \textbf{ours} & \textbf{0.9762}  & \textbf{0.9216} \\
        \bottomrule
    \end{tabular}
    \label{Tab: generative-face}
\end{table*}

\paragraph{Adversarial Attacks}
We adopt 5 gradient-based attacks, 3 adaptive attacks, 5 generative-based attacks, and 6 face attacks to evaluate our method.
For gradient-based attacks, BIM~\cite{BIM}, PGD~\cite{PGD}, MIM~\cite{MIFGSM}, DIM~\cite{DIFGSM}, and NIM~\cite{NIFGSM_SNIM} are used as the unseen attacks.
For adaptive attacks, we adopt AutoAttack (AA)~\cite{AutoAttack}, BPDA+EOT~\cite{BPDA_EBM}, and Minimum-Margin Attack (MM)~\cite{MM_Attack}.
For generative-based attacks, we adopt 3 GAN-based attacks including CDA~\cite{CDA}, TTP~\cite{TTP}, and M3D~\cite{M3D}, and 2 Diffusion-based attacks including Diff-Attack~\cite{Diff-attack} and Diff-PGD~\cite{Diff-pgd}.
For face attacks, we adopt 1 gradient-based attack TIPIM~\cite{TIPIM}, 2 GAN-based attacks including AdvMakeup~\cite{AdvMakeup} and AMT-GAN~\cite{AMT-GAN}, and 3 physical attacks including Adv-Sticker~\cite{advmask}, Adv-Glasses~\cite{advglass}, and Adv-Mask~\cite{advmask}.
For each of the above attack methods, adversarial data is generated with the attack intensity of $\epsilon = 4/255$ (except for the unrestricted attacks), and iterative attacks run for 10 steps using step size 2.

\paragraph{Baselines}
To evaluate our method across various types of adversarial attacks, we compare it with state-of-the-art detection methods from three related research domains: adversarial detection, synthetic image detection, and face forgery detection.
For gradient-based attacks, we benchmark against four adversarial detection methods: LID~\cite{LID}, LiBRe~\cite{Libre}, SPAD~\cite{Wang:23selfperturbation}, and EPSAD~\cite{EPSAD}.
Since most adversarial detection methods are not designed to address generative-based adversarial attacks, we include a comparison with four synthetic image detection techniques: CNN-Detection~\cite{cnn-detection}, LGrad~\cite{LGRAD}, Universal-Detector~\cite{universal-detector}, and DIRE~\cite{dire}.
For detecting GAN-based face attacks, we also compare against two face forgery detection methods, Luo et al.\cite{YuchenLuo2021} and He et al.\cite{YangHe2021}, as well as ODIN~\cite{ODIN}.

\paragraph{Evaluation Metrics}
Following previous adversarial detection works~\cite{EPSAD}, we employ the Area Under the Receiver Operating Characteristic Curve (AUROC)~\cite{AUROC} as the primary evaluation metric to assess the performance of our classification models. 
The AUROC is a widely recognized metric in the field of machine learning and is particularly useful for evaluating binary classification problems.
For fair comparisons, the number of natural data and adversarial data in the experiments remained consistent.

\paragraph{Implementation details}
XceptionNet~\cite{XceptionNet} is selected as our backbone model and is trained on all the above datasets.
For adversarial noise distribution estimation, we utilize Torchattacks~\cite{torchattacks} to generate 1000 noise images for each dataset.
The uniform distribution parameters are set as $\epsilon_{\mu} = 3$ and $\epsilon_{\sigma} = 0.005$.
The threshold of the sample control is set as $\gamma_s = 0.05$.
The thresholds of the binary indicator function are set as $\gamma_s = 0$, $\gamma_c = 125$ and $\gamma_l = 100$.
FGSM~\cite{FGSM} is used as the initial attack, i.e. the commonly used attack, for Perturbation Forgery.
Training epochs are set to 10 and convergence is witnessed. 

\subsection{Detecting Gradient-based Attacks}
\label{Sec: Exp-Gradient-based Attacks}

We begin by comparing our method with state-of-the-art techniques in detecting gradient-based attacks on ImageNet100. 
All the attacks are unseen by both the baseline methods and our method. 
For each attack, 1,000 adversarial images are generated in the gray-box setting, where the attacker has access to the victim model but is blind to the detection model. 
All baseline methods are sourced from existing adversarial detection research.

As reported in Table~\ref{Tab: ImageNet100-gradient}, although EPSAD performs well, our method demonstrates effective detection (achieving an AUROC score of 0.9931 against MIM and 0.9944 against MM) across a variety of attack methods. 
These results validate our assumptions about the proximity of the noise distribution.
By continuously perturbing the estimated noise distribution of a given attack (FGSM), we obtain an open covering noise of distributions from all the attacks.
Given one seen attack, we are able to detect all the unseen attacks.
The consistent detection performance verifies the effectiveness of our core idea of training a detector to distinguish the open covering of adv-noise distributions from natural ones and demonstrates the strong generalization capability of our method against gradient-based attacks. 

\subsection{Detecting Generative-based Attacks}
\label{Sec: Exp-generative-based Attacks}


To investigate our method's advantage in detecting generative-based attacks, we compared it with state-of-the-art synthetic image detection and adversarial detection methods on ImageNet100. As shown in Table~\ref{Tab: ImageNet100-generative}, our method significantly outperforms the baseline methods, demonstrating highly effective detection capabilities. 

For adversarial detection baselines, the imbalanced noise generated by GAN and diffusion models makes it difficult for these methods, which are typically designed for gradient-based attacks, to detect effectively. In contrast, with the help of sparse masks, our detector is able to detect the localized noise introduced by GANs and diffusion models. 

Most synthetic image detection methods are less effective in detecting adversarial data crafted from natural samples, with the exception of DIRE\cite{dire}. This effectiveness may stem from DIRE’s use of reconstruction error from images generated by diffusion models, which maintains its robustness in this setting.

\subsection{Detecting Face Adversarial Examples}
\label{Sec: Exp-face Attacks}

Adversarial attacks targeting face recognition models represent an important branch of adversarial example research.
Therefore, the detection ability of adversarial face examples is an important indicator of the performance of the detector.
We perform various adversarial face attacks and compare our approach with state-of-the-art methods from adversarial face detection and face forgery detection.

As shown in Table~\ref{Tab: generative-face}, our method shows excellent performance against a variety of adversarial attacks targeting faces and surpasses many existing adversarial face detection techniques. 
It further indicates the ability of the sparse mask to capture the essential characteristics of face adversarial examples and the strong generalization capability of our method.

\begin{table}[ht]
\footnotesize
    \centering
    \tabcolsep=0.19cm
    \caption{
    Ablation study on masks (AUROC score), where $f_{s}$ denotes saliency detection model and $f_{c}$ represents the CAM model.
    }
    \begin{tabular}{l|ccccc}
        \toprule
        Ablation & CDA & TTP & M3D & Diff-Attack & Diff-PGD    \\
        \hline
        w/o $f_{s}$        &  0.9755 & 0.9585 & 0.9320 & 0.9197 & 0.9135 \\
        w/o $f_{c}$        &  0.9864 & 0.9510 & 0.9262 & 0.9031 & 0.9062 \\
        w/o $\operatorname{Sobel}$        &  0.9868 & 0.7235 & 0.6984 & 0.5231 & 0.5547 \\
        \bottomrule
    \end{tabular}
    \label{Tab: ablation-mask}
\end{table}

\subsection{Ablation Study and Analysis}
\label{Sec: Ablation}

To further evaluate our method, we report an ablation study on the sparse mask, white-box attack detection results, the impact of the initial attack and the data size for distribution estimation, and the functionality of Perturbation Forgery through visualizations.
The impact of hyper-parameters and other analyses are reported in the supplementary materials.

\subsubsection{Impact of each mask}
As argued, the sparse mask is proposed to assist the detector in identifying local adversarial noises in generative-based adversarial data. 
As shown in Table~\ref{Tab: ablation-mask}, the absence of the saliency detection model $f_{s}$ or the GradCAM model $f_{c}$ has little effect on the overall system. 
This is because the regions masked by $f_{s}$ and $f_{c}$ are somewhat complementary. 
However, without the Sobel operation to sparsify the mask, the detector fails to detect most attacks, except for CDA. This is because generative-based adversarial noise tends to be imbalanced and localized, while CDA’s noise is more global. The generated pseudo-adversarial noise without Sobel becomes too dense, preventing the detector from effectively focusing on localized noise.

\subsubsection{Functionality Visualization}

\begin{figure}[t]
    \centering
    \includegraphics[width=0.98\columnwidth]{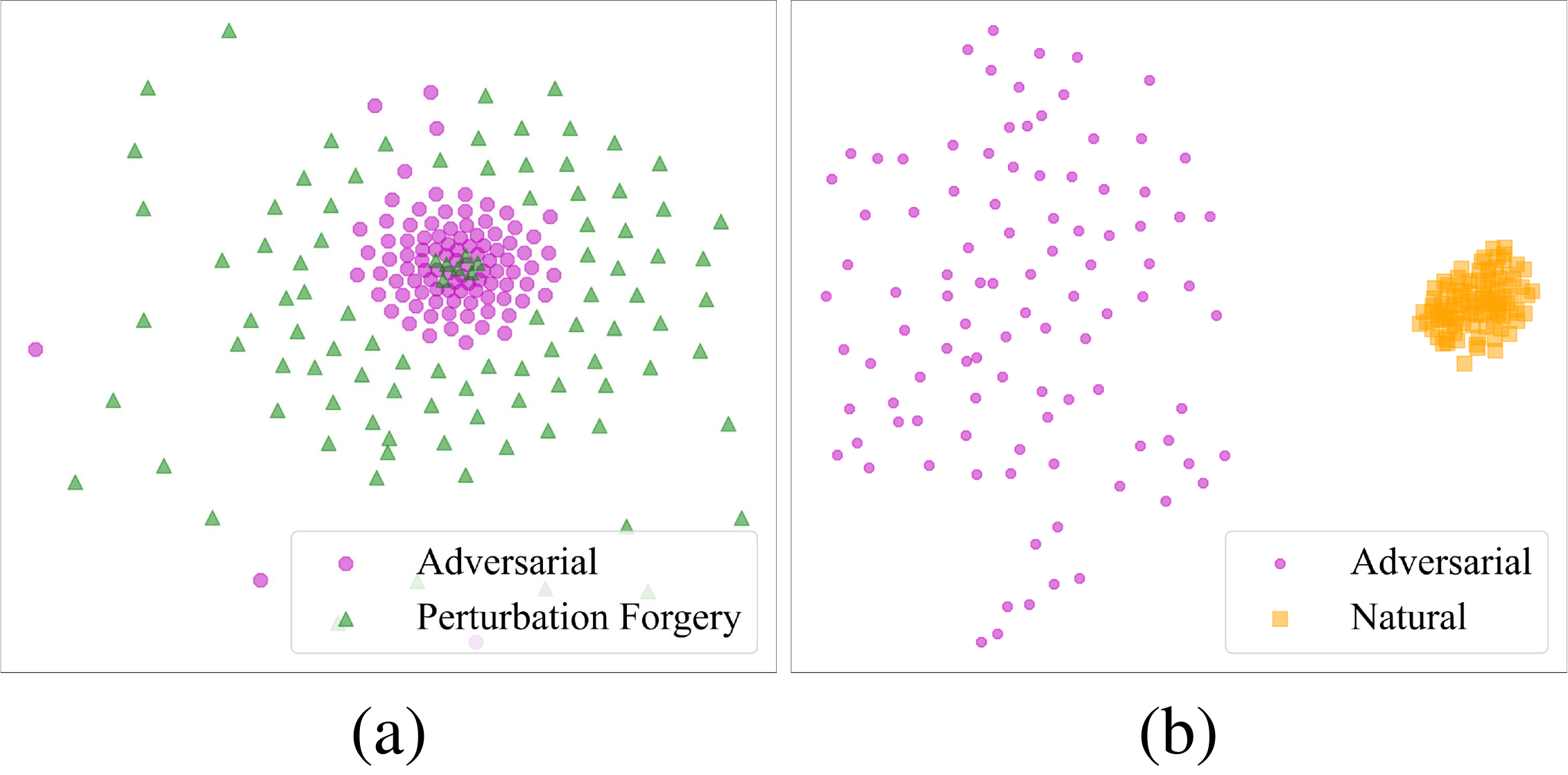}
    \caption{
    2D T-SNE visualizations.
    (a) ImageNet100 flattened noises of adversarial data and Perturbation Forgery-generated data.
    (b) ImageNet100 features extracted by the backbone model trained with Perturbation Forgery.
    }
    \label{fig: tsne}
\end{figure}

To validate the functionality of our method, we extract features and flattened noise from adversarial data and Perturbation Forgery-generated data, then visualize them using 2D t-SNE projections. 
All adversarial data originate from unseen attacks. 
As shown in Figure~\ref{fig: tsne}(a), Perturbation Forgery continuously perturbs the adv-noise distribution of the initial attack, thereby forming an open covering for noises from unseen attacks. 
Individual samples appear as outliers, floating outside the collective noise distribution. 
As shown in Figure~\ref{fig: tsne}(b), by learning to separate natural data from those generated by Perturbation Forgery, the trained detector can effectively distinguish between natural and adversarial data.

\begin{table}[t]
\footnotesize
    \centering
    \tabcolsep=0.275cm
    \caption{
    AUROC scores against white-box attacks under $ \epsilon= 4/255$ on CIFAR-10.
    }
    \begin{tabular}{l|ccccc}
        \toprule
        Attack & BIM & PGD & DIM & MIM & AA     \\
        \hline
        AUROC  &  0.9180 & 0.9168 & 0.9062 & 0.9031 & 0.9019 \\
        \bottomrule
    \end{tabular}
    \label{Tab: white-box}
\end{table}

\begin{table}[t]
\footnotesize
    \centering
    \tabcolsep=0.18cm
    \caption{
    Impact of data size on distribution estimation: AUROC score against PGD under $ \epsilon= 4/255$ on CIFAR-10.
    }
    \begin{tabular}{l|cccccc}
        \toprule
        Data size & 100 & 250 & 500 & 750 & 1000 & 2000     \\
        \hline
        AUROC  &  0.8278 & 0.9368 & 0.9866 & 0.9921 & 0.9965 & 0.9948\\
        \bottomrule
    \end{tabular}
    \label{Tab: dis-data size}
\end{table}

\subsubsection{Detecting White-box Attack}
To verify the robustness of our approach to white-box attacks, i.e., both the victim model and the detector are available to the attacker, we conduct the experiment on CIFAR-10. 
Following~\cite{NotEasyDetect}, 
a modified objective of attacks is:
\begin{equation}
    \max_{\boldsymbol{\eta}} \mathcal{L}(f(\hat{\boldsymbol{\mathrm{x}}}), y)+\alpha\cdot\mathcal{L}(f_d(\hat{\boldsymbol{\mathrm{x}}}), y_d), \ s.t. ||\boldsymbol{\eta}||\leq \epsilon
\end{equation}
where $y$ is the ground-truth label, $y_d$ = 0 indicates that $\hat{\boldsymbol{\mathrm{x}}}$ is an adversarial example, and $\alpha=1$ is a constant balancing between the attack of the classifier $f$ and detector $f_d$. 
As shown in Table~\ref{Tab: white-box}, our method achieves a high average AUROC of 0.9092, demonstrating its robustness against white-box attacks.
This may be attributed to the comprehensive open covering formed by perturbation forgery, which effectively encompasses adversarial data from white-box attacks, making it harder to mislead the trained detector.

\begin{figure}[ht]
    \centering
    \includegraphics[width=0.9\columnwidth]{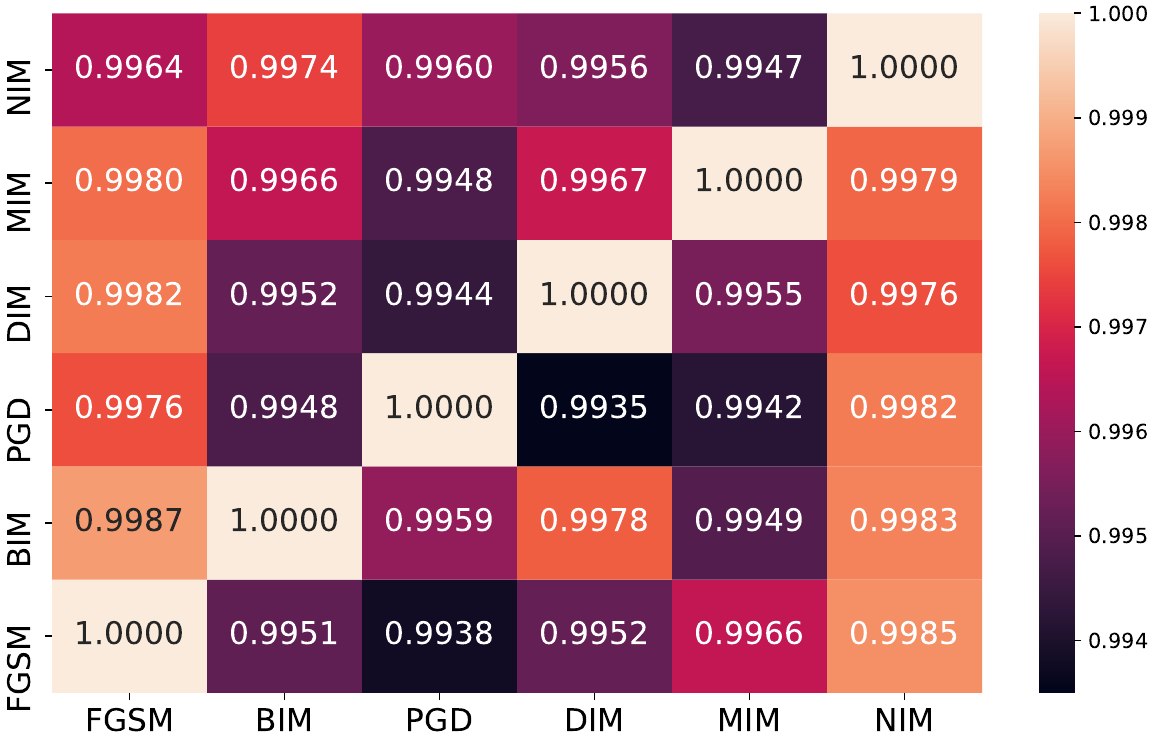}
    \caption{
    Impact of the initial attacks (AUROC) on CIFAR-10 under $\epsilon=4/255$. Y-axis: initial attack. X-axis: testing attack.
    }
    \label{fig: dis}
\end{figure}

\subsubsection{Impact of the Initial Attack}
Our proposed perturbation forgery requires a commonly used attack to construct the open covering. 
To assess the impact of the initial attack choice, we train detectors using distributions estimated from various initial attacks and test them against other attacks on CIFAR-10. 
As shown in Figure~\ref{fig: dis}, the detectors consistently achieve a high AUROC score above 0.99, indicating that our method does not depend on a specific attack.

\subsubsection{Data size for Distribution Estmation}

To assess the influence of data size for distribution estimation, we train detectors using distributions estimated from various data sizes and test them against PGD on CIFAR-10. 
As shown in Table~\ref{Tab: dis-data size}, our method only requires a small amount of data ($\geq 500$ samples) to estimate the adversarial noise distribution so as to train an effective detector.
Notably, using fewer than 500 samples leads to a significant decrease in detection performance.

\section{Conclusion}
\label{Sec: Conclusion}

In this paper, we explore and define the proximity relationship between adversarial noise distributions by modeling these distributions as multivariate Gaussian. 
Based on our assumption and definition, we establish a perturbation method to craft proximal distributions from a given distribution and demonstrate the existence of an open covering of adversarial noise distributions. 
After that, we propose the core idea of obtaining a detector with strong generalization abilities against various types of adversarial attacks by training on the open covering of adversarial noise distributions.
Based on this insight, we heuristically propose Perturbation Forgery, which includes noise distribution perturbation, sparse mask generation, and pseudo-adversarial data production, to train an adversarial detector capable of detecting unseen gradient-based, generative-based, and physical adversarial attacks, agnostic to any specific models.
Extensive experiments confirm the effectiveness of our approach and demonstrate its strong generalization ability against unseen adversarial attacks, with a satisfactory inference time overhead.


{
    \small
    \bibliographystyle{ieeenat_fullname}
    \bibliography{PFD}
}


\end{document}